%% file: main.tex
\documentclass[10pt,twocolumn,letterpaper]{article}

\usepackage[pagenumbers]{cvpr}   
\usepackage{graphicx}
\usepackage{capt-of} 
\usepackage{multirow}
\usepackage{textcomp}
\usepackage{booktabs}
\usepackage{amssymb}
\usepackage{pifont}
\usepackage{array}
\usepackage{float}


\definecolor{cvprblue}{rgb}{0.21,0.49,0.74}
\usepackage[
    pagebackref,
    breaklinks,
    colorlinks,
    allcolors=cvprblue
]{hyperref}

\def\confName{CVPR}
\def\confYear{2026}

\title{PixDLM: A Dual-Path Multimodal Language Model for UAV Reasoning Segmentation}

\author{
Shuyan Ke$^{1}$, Yifan Mei$^{1}$, Changli Wu$^{\dagger1,2}$, Yonghan Zheng$^{1}$, Jiayi Ji$^{1}$, Liujuan Cao$^{1}$, Rongrong Ji$^{\ast1}$\\
$^{1}$ Key Laboratory of Multimedia Trusted Perception and Efficient Computing,\\
Ministry of Education of China, Xiamen University, 361005, P.R. China\\
$^{2}$ Shanghai Innovation Institute\\
{\tt\small \{keshuyan, meiyifan, wuchangli, zhengyonghan\}@stu.xmu.edu.cn}\\
{\tt\small jjyxmu@gmail.com \quad \{caoliujuan, rrji\}@xmu.edu.cn}
}

\begin{document}

\twocolumn[{%
\renewcommand\twocolumn[1][]{#1}%
\maketitle 

\begin{center}
\vspace{-0.5cm}
\includegraphics[width=0.85\textwidth]{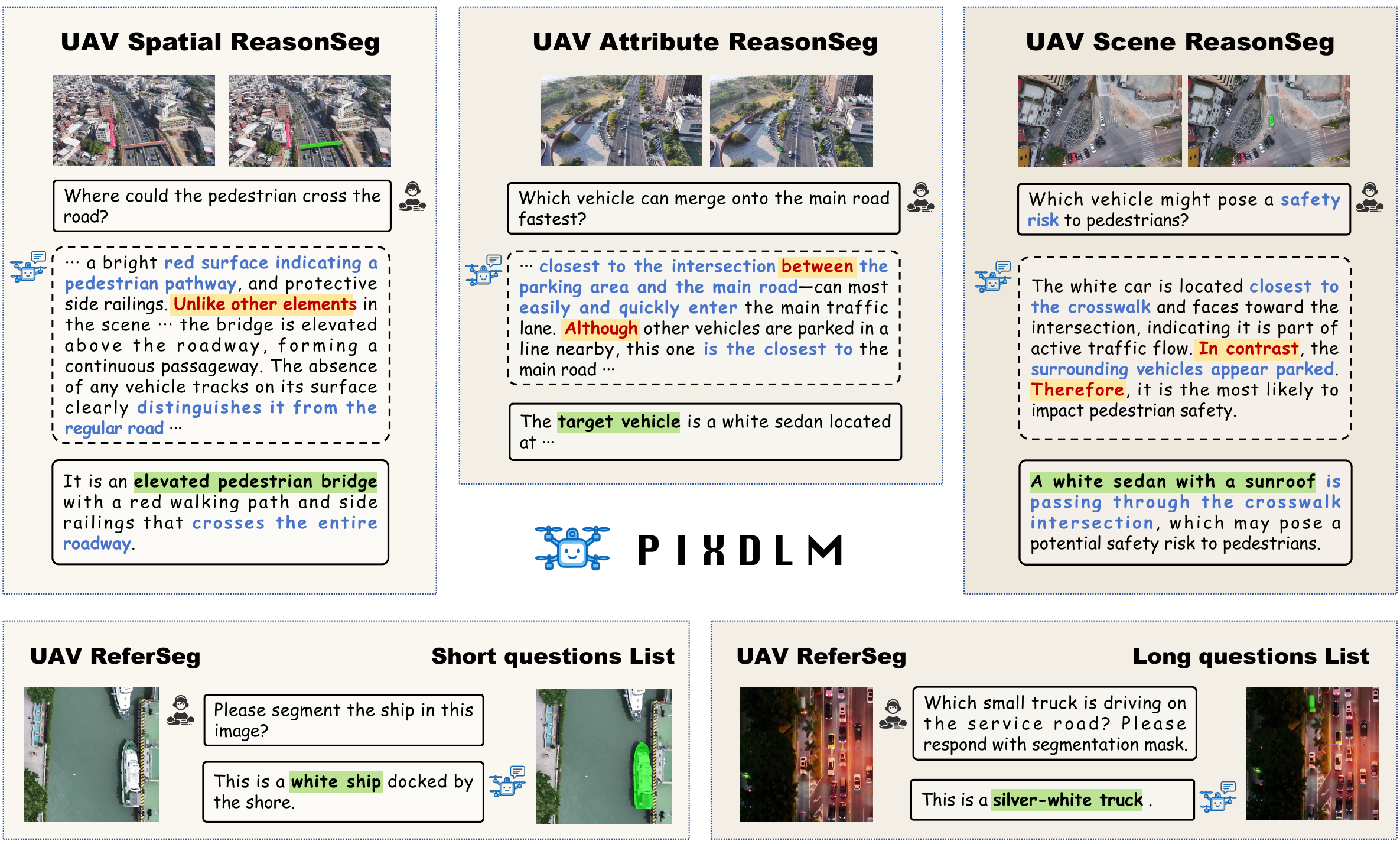}
\captionof{figure}{
Overview of the UAV ReasonSeg task, which involves three core reasoning types: spatial, attribute, and scene-level reasoning, illustrated with UAV image pairs, questions, and reasoning traces. 
The bottom row presents UAV referring segmentation under both short and long language instructions.
}
\label{fig:1}
\end{center}
}]

\let\thefootnote\relax
\footnotetext{$^{\ast}$Corresponding author $^{\dagger}$Project leader}

\input{sec/0_abstract}

\input{sec/1_intro}
\input{sec/2_related}

\input{sec/3_DRSeg}

\input{sec/4_method}

\input{sec/5_experiment}

\input{sec/6_conclusion}
\input{sec/7_acknowledgments}
\input{sec/9_supple}

{
    \small
    \bibliographystyle{ieeenat_fullname}
    \bibliography{main}
}

\end{document}

%% file: sec/0_abstract.tex
\begin{abstract} 
Reasoning segmentation has recently expanded from ground-level scenes to remote-sensing imagery, yet UAV data poses distinct challenges, including oblique viewpoints, ultra-high resolutions, and extreme scale variations. To address these issues, we formally define the UAV Reasoning Segmentation task and organize its semantic requirements into three dimensions: Spatial, Attribute, and Scene-level reasoning. Based on this formulation, we construct DRSeg, a large-scale benchmark for UAV reasoning segmentation, containing 10k high-resolution aerial images paired with Chain-of-Thought QA supervision across all three reasoning types. As a benchmark companion, we introduce PixDLM, a simple yet effective pixel-level multimodal language model that serves as a unified baseline for this task. Experiments on DRSeg establish strong baseline results and highlight the unique challenges of UAV reasoning segmentation, providing a solid foundation for future research.
All datasets, models, and code are available at \url{https://github.com/XIEFOX/PixDLM}.
\end{abstract}

%% file: sec/1_intro.tex
\section{Introduction}
\label{sec:intro}
\begin{figure*}[t]
    \centering
    \includegraphics[width=1\linewidth]{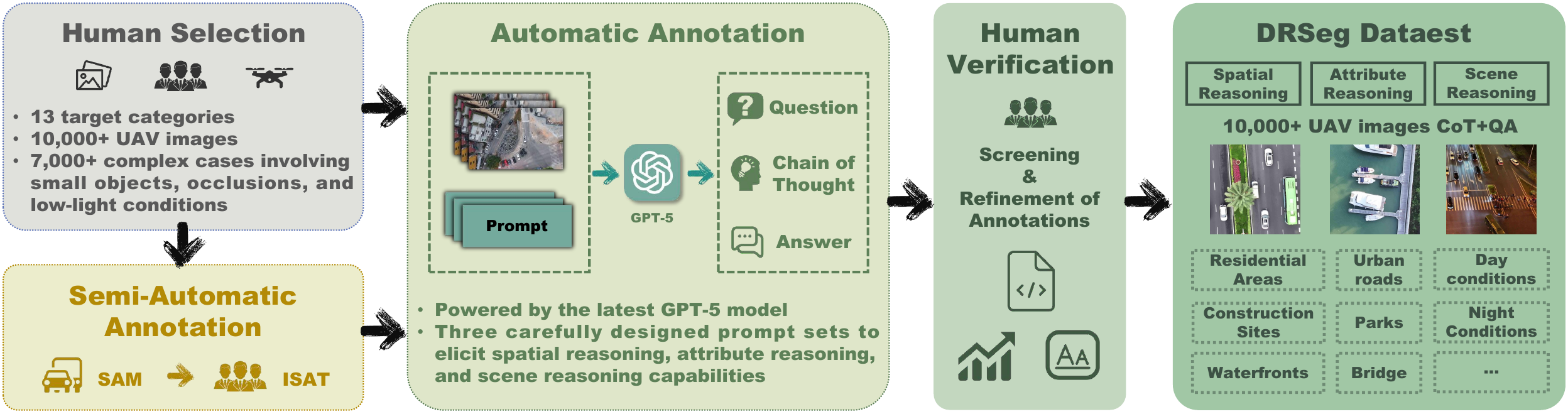}

   \caption{Overview of the DRSeg construction pipeline. Includes (1) human selection, (2) semi-automatic mask generation, (3) automatic reasoning annotation, and (4) human verification, producing a high-quality UAV reasoning segmentation dataset with multi-level supervision.
}

    \label{fig:con}
\end{figure*}

Reasoning segmentation aims to identify image regions that satisfy free-form natural language instructions~\cite{LaiTCLY0J24,RenHW0FFJ24,liu2023visual,abs-2409-12191,liu2023interngpt,liu2024primitivenet}. 
Recent works such as LISA~\cite{LaiTCLY0J24} and PixelLM~\cite{RenHW0FFJ24} show that large multimodal models can perform implicit and relation-based reasoning for pixel-level segmentation, establishing reasoning segmentation as an instruction-driven visual understanding paradigm.
However, existing reasoning segmentation benchmarks~\cite{ChengYYYZXH23,XiaBDZBLDPZ18,ChengWLXLYH22,LiuYWY17,abs-2003-06832,RazakarivonyJ16}  and models are largely built on \textit{ground-view} or \textit{nadir-view} imagery~\cite{abs-2306-15195,0008LSH23}. 
Their visual assumptions moderate resolution, limited scale variation, stable camera orientation, and sizable objects—do not hold for unmanned aerial vehicle (UAV) imagery. 
In particular, UAV imagery presents three defining characteristics: 
(i) \textbf{High-altitude and oblique viewpoints} continuously change the perspective geometry, requiring models to infer semantics across dynamic view changes rather than from a fixed ground-level reference frame. 
(ii) \textbf{Extreme scale variation and dense small objects} mean that many critical targets---vehicles, pedestrians, debris, and inspection structures---occupy only tens of pixels, making them easily lost during token compression in existing MLLMs. 
(iii) \textbf{Ultra-high-resolution scenes} combine global environmental context with fine-grained structural cues (e.g., rooftops, lanes, and fences), requiring models to reason jointly over global semantics and tiny, high-frequency details.
We therefore define the \textbf{UAV Reasoning Segmentation} task(Fig.~\ref{fig:1}), which takes a UAV image and a free-form instruction as input and predicts a pixel-level mask for the region that satisfies the instruction. 
Unlike referring segmentation, which often benefits from explicit object mentions~\cite{abs-2501-06828,abs-2501-13925,LiuMZ0JSJ24}, UAV instructions typically emphasize \textit{spatial configuration} (``the vehicle closest to the crosswalk''), \textit{visual state} (``the solar panel with broken cells''), or \textit{scene intent} (``the area suitable for landing''). 

Despite its practical importance in low-altitude monitoring, search-and-rescue, intelligent inspection, and autonomous drone navigation, reasoning segmentation under UAV conditions still  \emph{lacks a dedicated benchmark} because existing UAV datasets generally lack fine-grained annotations, diverse semantic coverage, and reasoning-oriented textual supervision~\cite{abs-2507-12883,0007ZLLKKKD25}.
Several UAV-based datasets such as VisDrone~\cite{DuZBSZHZLWWWZFC19}, AUAIR~\cite{BozcanK20}, UAVDT~\cite{DuQYYDLZHT18}, and CARPK~\cite{NascimentoCKGFB20} provide drone-view images but generally lack fine-grained or oriented annotations. 
Even datasets specifically designed for UAV-oriented detection, including DroneVehicle~\cite{SunCZH22} and UAV-ROD~\cite{ZhouFLHP22}, cover only a narrow set of object categories and do not include reasoning-oriented textual instructions~\cite{SunGMZFLY24,WangSWWG22}, making them inadequate for tasks that require diverse scene elements and instruction-dependent target definitions.
This gap prevents the community from systematically measuring progress in UAV reasoning and limits the development of models capable of supporting real-world aerial decision-making.

To address this gap, we introduce \textbf{DRSeg}, a benchmark tailored to UAV Reasoning Segmentation. 
DRSeg contains over 10,000 high-resolution UAV images across diverse environments and illumination conditions, enriched with pixel-accurate masks and Chain-of-Thought reasoning QA pairs reflecting three key reasoning dimensions—spatial, attribute, and scene-level reasoning. 
Each QA–mask pair is generated through a GPT-5 guided process and validated by human experts, ensuring consistency between reasoning intention and final pixel-level supervision.

However, a benchmark alone is insufficient to facilitate systematic evaluation of UAV reasoning segmentation. Existing reasoning segmentation models typically rely on low-resolution visual tokenization, which may discard fine-grained UAV details during compression. To provide a reference baseline under such conditions, we introduce PixDLM, a pixel-level multimodal language model equipped with a Dual-Path Vision Encoder. The model combines a global low-resolution pathway for semantic context with a high-resolution structural pathway that preserves small objects and boundary cues. These pathways are integrated through a lightweight MultiPath Alignment module, while a hierarchical decoder refines segmentation masks under LLM-guided reasoning. Our contributions are threefold:
\begin{itemize}
    \item We \textbf{formalize UAV Reasoning Segmentation} as an instruction-driven pixel-level prediction task under UAV-specific visual statistics, clarifying why existing reasoning models fail under aerial viewpoints.
    \item We introduce \textbf{DRSeg}, a large-scale UAV reasoning segmentation benchmark with over 10,000 high-resolution images and CoT-aligned reasoning annotations.
    \item We introduce PixDLM, a Dual-Path pixel-level multimodal language model that serves as a strong baseline for UAV reasoning segmentation and demonstrates competitive performance on standard referring segmentation benchmarks.
\end{itemize}

%% file: sec/2_related.tex
\section{Related Work}
\label{sec:Related}

\paragraph{Large Multimodal Models.}
Recent LMMs such as BLIP-2~\cite{0008LSH23}, InternGPT~\cite{liu2023interngpt}, LLaVA~\cite{liu2023visual}, Llama-Adapter~\cite{ZhangHLZL00024}, and Qwen-VL~\cite{abs-2409-12191} have significantly advanced open-world visual reasoning by aligning vision and language through adapter-based fusion and instruction tuning.  
However, these models primarily produce text-only outputs and lack pixel-level prediction mechanisms, making them unsuitable for fine-grained spatial reasoning—especially under UAV conditions involving oblique viewpoints, high-resolution imagery, and extreme scale variation.

\paragraph{Reasoning Segmentation.}
LISA~\cite{LaiTCLY0J24} formalized reasoning segmentation by introducing a multimodal LLM with a $<\text{SEG}>$ token and an embedding-as-mask decoder, enabling implicit language reasoning with pixel-level outputs.  
PixelLM~\cite{RenHW0FFJ24} further removed external segmentation modules by integrating a lightweight pixel decoder and segmentation codebook, supporting multi-object and multi-scale reasoning.  
While effective on ground-view scenes, these models generalize poorly to UAV imagery~\cite{abs-2503-06520,WangK22,abs-2312-17240,abs-2407-14500}, where viewpoint distortions, tiny objects, and high-frequency structural cues are easily lost through low-resolution tokenization~\cite{liu2025boost}.

\paragraph{Reasoning in Geographic and Remote Sensing Domains.}
In recent years, deep learning has made remarkable progress in the field of remote sensing, significantly advancing the ability to interpret large-scale aerial and satellite imagery~\cite{liu2025boost,yao2025remotesam,yao2025remotereasoner,liu2024remoteclip}. 
Building on this momentum, the reasoning paradigm has been extended to large-scale geographic scenes, with GeoPix~\cite{abs-2501-06828} and GeoPixel~\cite{abs-2501-13925} incorporating spatial priors and topological cues to improve language-guided mask generation.  
Meanwhile, remote-sensing models such as RS-MMFormer~\cite{ZhangMLFHZ23}, RemoteSAM~\cite{abs-2505-18022}, and AerialVLN\cite{LiuZQ0ZW23} adapt generic segmentation frameworks to high-resolution satellite imagery.  
Nonetheless, these approaches remain bound to predefined semantic categories and show limited open-vocabulary reasoning, weak compositionality, and insufficient robustness to dense small-object distribution~\cite{CaoCMY25,LiQ25,abs-2410-17283,yao2025remotereasoner,liu2024remoteclip}.

\paragraph{Challenges in UAV Imagery.}
Compared with satellite or ground-view images, UAV scenes feature drastic viewpoint changes, complex motion patterns, and dynamic scene characteristics, which motivate the development of DRSeg as a standardized benchmark for language-driven reasoning in aerial imagery~\cite{abs-1804-07437,YuLZHDTS20,verykokou2018oblique}~\cite{abs-1804-07437,YuLZHDTS20,verykokou2018oblique}.  
Moreover, Existing reasoning segmentation models fail to preserve small, densely distributed targets, while remote-sensing frameworks do not generalize well to low-altitude oblique perspectives and often incur prohibitive computational costs~\cite{XiaBDZBLDPZ18,li2020object}. Inspired by LLaVA-HR~\cite{luo2024feast}, PixDLM introduces a dual-path visual encoder that explicitly balances global semantic abstraction and high-resolution structural perception through controlled integration.

%% file: sec/3_DRSeg.tex
\begin{figure*}[t]
    \centering
    \includegraphics[width=1\linewidth]{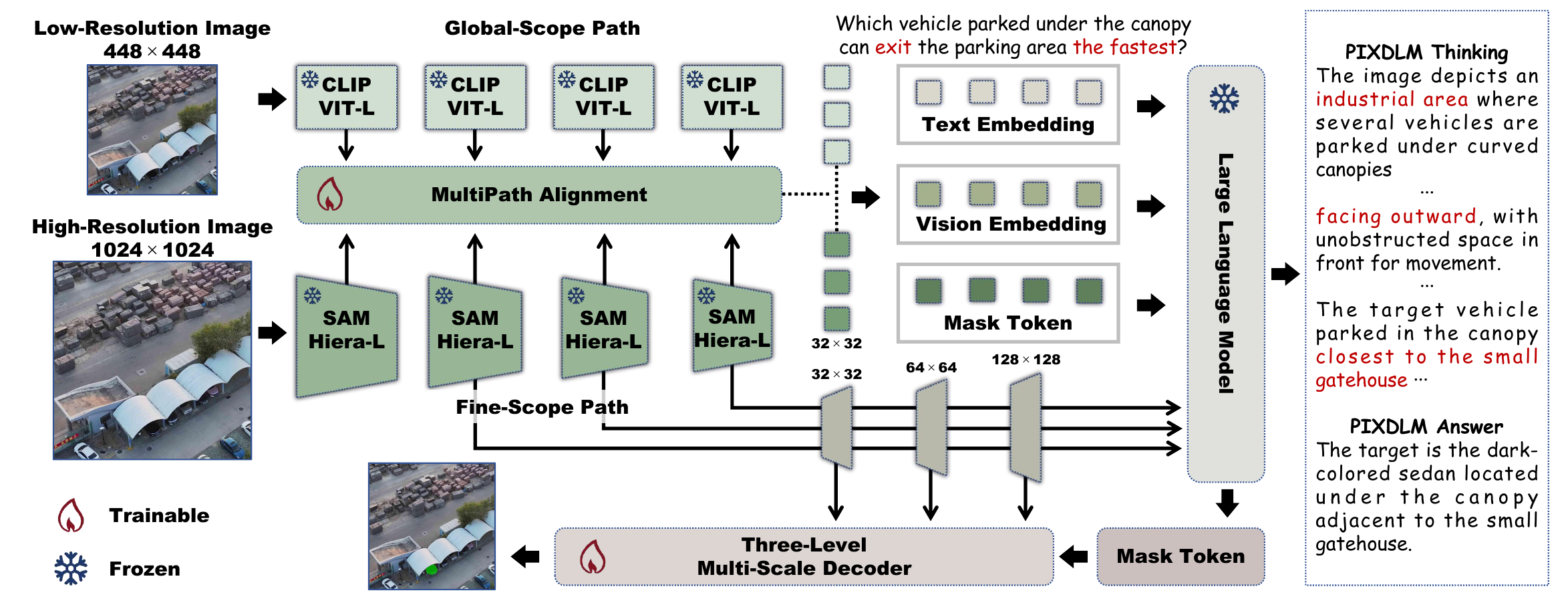}
    \caption{
Overview of PixDLM. A Dual-Path Vision Encoder extracts global semantics (CLIP) and fine-grained structure (SAM), which are fused through MultiPath Alignment. Text embeddings and a Mask Token are processed by the LLM for instruction-conditioned reasoning, and a multi-scale decoder reconstructs the final segmentation mask.
}
    \label{fig:method}
\end{figure*}

\section{Task Formulation and Dataset Construction}

\subsection{Task Formulation}
Given an input UAV image \(x_{\text{img}}\) and an implicit natural language instruction \(x_{\text{txt}}\), the goal of \textbf{UAV Reasoning Segmentation} is to predict a binary mask \(M \in \{0,1\}^{H \times W}\) satisfying the semantic conditions expressed in the instruction:
\begin{equation}
    M = f_{\theta}(x_{\text{img}}, x_{\text{txt}}),
\end{equation}
where \(f_{\theta}\) denotes a multimodal reasoning model parameterized by~\(\theta\).

Unlike conventional referring segmentation, UAV ReasonSeg operates in a \textbf{category-agnostic} and \textbf{implicit} setting. Instructions do not explicitly name target categories (e.g., ``the red truck''), but instead describe relational, contextual, or intent-driven semantics such as ``the safest open area for landing'', ``the object blocking the road ahead'', or ``the building with visible structural damage''. This requires models to infer \emph{what} to segment from scene context and reasoning cues rather than surface-level category recognition.

To capture the diversity of implicit semantics in UAV scenes, we define three complementary reasoning dimensions:
\begin{itemize}
    \item \textbf{Spatial Reasoning}: inferring geometric or positional relations (e.g., ``the vehicle behind the fire truck'', ``the region closest to the riverbank'').
    \item \textbf{Attribute Reasoning}: identifying latent visual properties or object states (e.g., ``the illuminated structure at night'', ``the damaged solar panel'').
    \item \textbf{Scene-level Reasoning}: interpreting global context or task-oriented semantics (e.g., ``the safest landing zone'', ``the area suggesting ongoing construction'').
\end{itemize}

Compared with ground-view or satellite-view imagery, UAV images exhibit unique challenges including oblique viewpoints, large-scale variation, dense small objects, and dynamic flight conditions. Existing reasoning segmentation datasets do not reflect these characteristics, motivating the need for a UAV-specific benchmark.

We evaluate performance using two metrics: gIoU and cIoU. Specifically, the generalized Intersection-over-Union (gIoU) measures the average IoU across individual images. In contrast, the cumulative Intersection-over-Union (cIoU) computes the ratio between the total intersection and total union aggregated over the entire dataset.

\subsection{DRSeg Dataset Construction}
To fill this gap, we present \textbf{DRSeg}, the first large-scale dataset tailored for UAV reasoning segmentation. Built upon the CODrone dataset~\cite{abs-2504-20032,abs-2507-20920}, DRSeg spans diverse aerial environments such as urban roads, parks, residential areas, industrial zones, and waterfront scenes. Images are captured from multiple altitudes (30\,m/60\,m/100\,m) and various lighting conditions, including nighttime and mixed artificial illumination.

The construction pipeline of DRSeg consists of four stages (Fig.~\ref{fig:con}): 

\textbf{Human Selection.}
We first curate UAV images exhibiting strong visual and semantic complexity, including large scale variation, dense objects, motion blur, occlusion, and oblique viewpoints. Low-quality or trivial images are filtered out to ensure challenging and diverse reasoning scenarios. 

\textbf{Semi-Automatic Annotation.}
Rotated bounding boxes from CODrone are converted into coarse instance masks via SAM2~\cite{KirillovMRMRGXW23}. These masks are further refined using the semi-automatic ISAT tool to correct boundary errors, enabling accurate pixel-level instance segmentation, especially for small objects or thin structures.

\textbf{Automatic Reasoning Annotation.}
To generate reasoning-aware supervision, we employ GPT-5 conditioned on the image, refined mask, and category label. Three tailored prompt sets (See in the Supplementary Material) guide GPT-5 to produce reasoning-oriented question--answer pairs across spatial, attribute, and scene-level dimensions. Each annotation includes natural language reasoning and a distilled Chain-of-Thought (CoT) trace, giving interpretable and context-rich supervision. We conduct masking, shuffling, and semantic corruption experiments on the CoT traces of 10\% of the samples, and the limited performance changes indicate strong robustness to noisy reasoning supervision.

\textbf{Human Verification.}
Finally, human annotators review all automatically generated annotations for logical consistency, correctness, and semantic alignment with the target mask. Valid samples are retained and split into training, validation, and test sets following a 3:2:5 ratio.

\subsection{Dataset Scale and Statistics}
DRSeg contains a total of \textbf{10,000 high-resolution UAV images} and \textbf{10,000 instance masks}, making it one of the largest UAV-view segmentation datasets to date.  
Although each image is annotated with only one target instance for reasoning, aerial scenes typically contain many distractor objects, leading to high effective object density.

For reasoning supervision, DRSeg provides \textbf{10,000 reasoning QA pairs}, one for each annotated instance, covering all three reasoning dimensions.
Among them, \textbf{33.33\%} correspond to Spatial reasoning, \textbf{33.34\%} to Attribute reasoning, and \textbf{33.33\%} to Scene-level reasoning, demonstrating a balanced distribution that supports diverse reasoning behaviors.

In terms of geometric diversity, the dataset spans three flight altitudes (30\,m, 60\,m, and 100\,m), with \textbf{31.44\%}, \textbf{25.45\%}, and \textbf{43.11\%} of images captured at each altitude, respectively.  
Object scale variation is equally significant: \textbf{58.08\%} of all instances fall into the small-object regime (area $<\! 2\%$ of the image), highlighting the fine-grained challenges introduced by UAV viewpoints.Additional dataset visualizations are provided in the Supplementary Material.

In total, DRSeg provides a high-resolution, reasoning-rich benchmark specifically aligned with the challenges of UAV perception, enabling systematic evaluation of multimodal reasoning and segmentation models.

%% file: sec/4_method.tex
\begin{table*}[t]
\centering
\caption{Comparison of different models on the \textbf{DRSeg} benchmark across three reasoning dimensions.}
\large  
\setlength{\tabcolsep}{6pt}
\renewcommand{\arraystretch}{1.15}
\resizebox{0.85\textwidth}{!}{
\begin{tabular}{l|l|cc|cc|cc}
\hline
\multirow{2}{*}{\textbf{Setting}} &
\multirow{2}{*}{\textbf{Model Name}} &
\multicolumn{2}{c|}{\textbf{Attribute Reasoning}} &
\multicolumn{2}{c|}{\textbf{Scene Reasoning}} &
\multicolumn{2}{c}{\textbf{Spatial Reasoning}} \\ 
\cline{3-8}
 & & $gIoU\uparrow$ & $cIoU\uparrow$ & $gIoU\uparrow$ & $cIoU\uparrow$ & $gIoU\uparrow$ & $cIoU\uparrow$ \\ 
\hline
\multirow{7}{*}{Zero-shot}
& SegEarth-R1~\cite{abs-2504-09644}         & 20.58 & 11.18 & 26.33 & 16.11 & 25.83 & 14.87 \\
& LISA-13B-llama2-v1~\cite{LaiTCLY0J24}     & 52.65 & 48.12 & 47.08 & 42.41 & 42.85 & 37.94 \\
& LISA-7B-v1                                & 49.21 & 45.15 & 45.55 & 41.59 & 42.00 & 37.50 \\
& LISA-7B-v1-explanatory                    & 48.40 & 43.77 & 42.79 & 37.86 & 38.69 & 34.97 \\
& LISAt-7B                                   & 43.80 & 46.36 & 42.13 & 44.20 & 39.91 & 37.45 \\
& PixelLM-7B~\cite{RenHW0FFJ24}             & 46.87 & 49.55 & 43.07 & 45.49 & 41.28 & 40.94 \\
& PixelLM-13B                                & 48.39 & 47.26 & 45.06 & 48.90 & 41.98 & 39.31 \\
\hline
\multirow{3}{*}{SFT}
& LISA-7B-v1                         & \underline{59.22} & \underline{59.66} & 54.45 & 53.25 & \underline{57.33} & \underline{58.02} \\
& PixelLM-7B                             & 58.39 & 57.35 & \underline{55.19} & \underline{54.29} & 55.38 & 54.47 \\
& \textbf{Ours}                              & \textbf{62.80} & \textbf{62.84} & \textbf{61.75} & \textbf{64.03} & \textbf{62.51} & \textbf{62.80} \\
\hline
\end{tabular}}
\label{tab:drseg_results}
\end{table*}

\section{Method}

While DRSeg enables systematic evaluation of UAV reasoning segmentation, effective benchmarking also requires a suitable baseline. Existing MLLMs typically rely on low-resolution visual tokens~\cite{0008LSH23}, which discard fine-grained UAV details, while directly reasoning over full-resolution imagery is computationally prohibitive due to substantial memory overhead. To provide a reference point for future research, we introduce \textbf{PixDLM}, a lightweight pixel-level multimodal model designed as a baseline for DRSeg.

\begin{figure*}[t]
    \centering
    \includegraphics[width=0.85\linewidth]{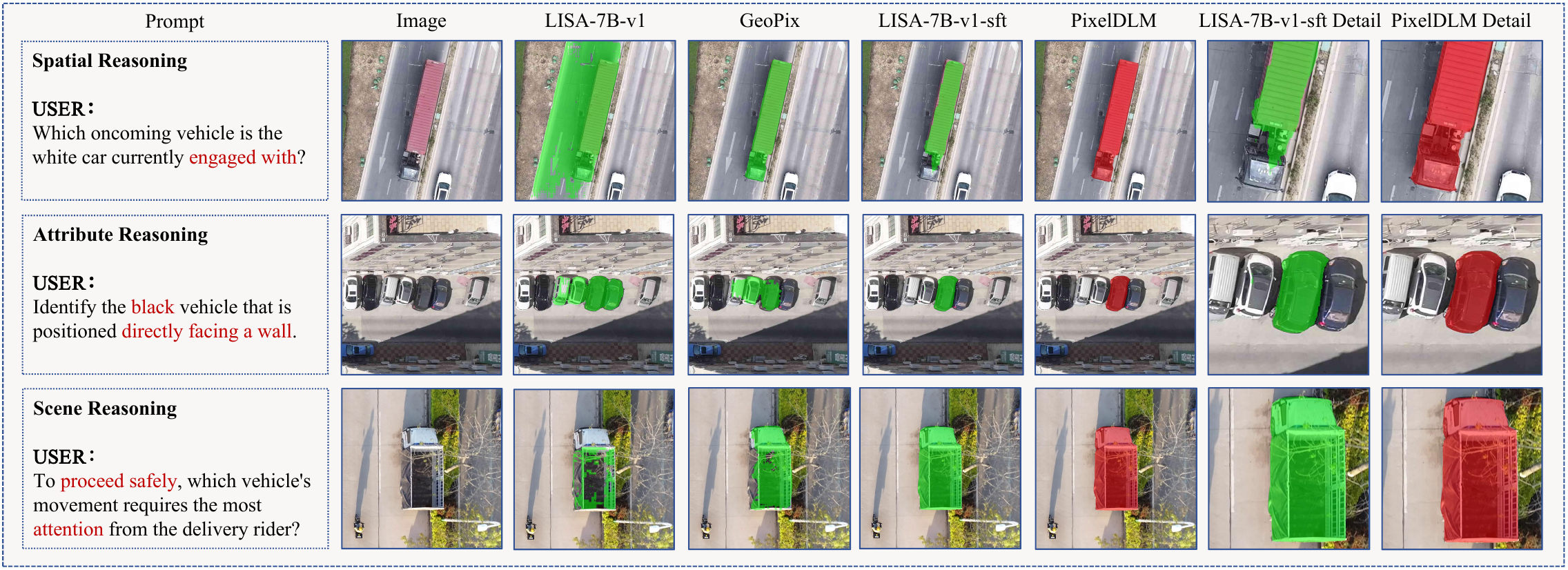}
    \caption{Visual comparison of segmentation results produced by LISA, our fine-tuned LISA variant, GeoPix, and our PixDLM model.
}

    \label{fig:comparison}
\end{figure*}

\begin{figure*}[t]
    \centering
    \includegraphics[width=0.80\linewidth]{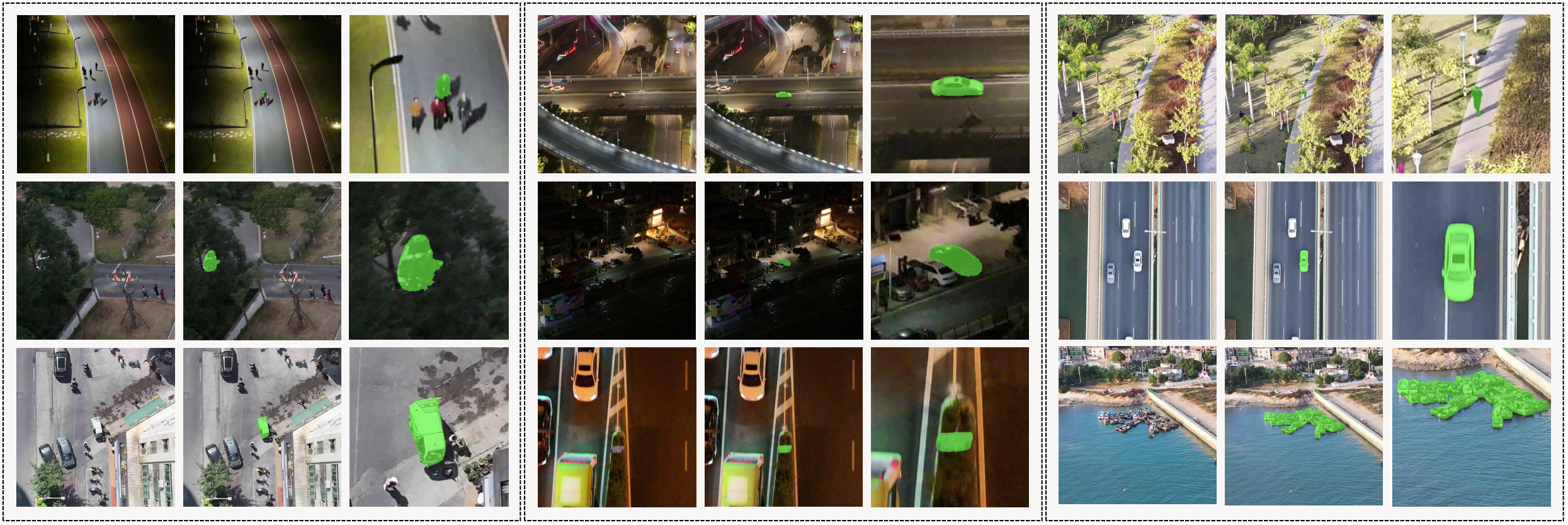}
    \caption{Visualization of PixDLM results on the three DRSeg reasoning subsets: Attribute (left), Scene (middle), and Spatial (right). 
Each panel shows examples with increasing reasoning complexity, diverse visual conditions, and varying localization difficulty.}
    \label{fig:segmentation results}
\end{figure*}
\begin{table*}[t]
\centering
\caption{
Ablation study on the effects of training datasets (\textbf{RRSIS-D}~\cite{LiuMZ0JSJ24}, \textbf{DRSeg}) and \textbf{COT} supervision. 
Each checkmark ($\checkmark$) indicates a 1:1 inclusion. 
We evaluate how different dataset combinations and COT usage influence reasoning performance. 
Metrics (\%) are reported as $gIoU\uparrow$ and $cIoU\uparrow$.
}
\setlength{\tabcolsep}{6pt}
\renewcommand{\arraystretch}{1.15}
\resizebox{0.80\textwidth}{!}{
\begin{tabular}{ccc|cc|cc|cc}
\hline
\multicolumn{3}{c|}{\textbf{Activated Datasets}} &
\multicolumn{2}{c|}{\textbf{Attribute Reasoning}} &
\multicolumn{2}{c|}{\textbf{Scene Reasoning}} &
\multicolumn{2}{c}{\textbf{Spatial Reasoning}} \\ 
\cline{4-9}
\textbf{COT} & \textbf{RRSISD} & \textbf{DRSeg} &
$gIoU\uparrow$ & $cIoU\uparrow$ &
$gIoU\uparrow$ & $cIoU\uparrow$ &
$gIoU\uparrow$ & $cIoU\uparrow$ \\
\hline
\checkmark & \checkmark & \checkmark & 61.13 & 60.60 & 55.60 & 54.53 & \underline{60.55} & \textbf{65.73} \\
           & \checkmark & \checkmark & 59.02 & 60.19 & \underline{55.98} & \underline{54.70} & 59.00 & 59.21 \\
\checkmark &            & \checkmark & \textbf{62.80} & \textbf{62.84} & \textbf{61.75} & \textbf{64.03} & \textbf{62.51} & \underline{62.80} \\
           &            & \checkmark & \underline{62.51} & \underline{61.62} & 55.92 & 54.29 & 59.57 & 60.94 \\
\hline
\end{tabular}}
\label{tab:cot_ablation}
\end{table*}

\begin{table*}[t]
\centering
\caption{Ablation study of the \textbf{Dual-Path Vision Encoder} on the \textbf{DRSeg} benchmark. 
Each layer is progressively activated from top to bottom. Metrics (\%) are reported as $gIoU\uparrow$ and $cIoU\uparrow$.}
\setlength{\tabcolsep}{6pt}
\renewcommand{\arraystretch}{1.15}
\resizebox{0.80\textwidth}{!}{
\begin{tabular}{cccc|cc|cc|cc}
\hline
\multicolumn{4}{c|}{\textbf{Activated Layers}} &
\multicolumn{2}{c|}{\textbf{Attribute Reasoning}} &
\multicolumn{2}{c|}{\textbf{Scene Reasoning}} &
\multicolumn{2}{c}{\textbf{Spatial Reasoning}} \\ 
\cline{5-10}
\textbf{1st} & \textbf{2nd} & \textbf{3rd} & \textbf{4th} &
$gIoU\uparrow$ & $cIoU\uparrow$ &
$gIoU\uparrow$ & $cIoU\uparrow$ &
$gIoU\uparrow$ & $cIoU\uparrow$ \\
\hline
 &   &   &   & 58.39 & 57.35 & 55.19 & 54.29 & 55.38 & 54.47 \\
  &  &   & \checkmark  & 58.90 & 59.77 & 55.38 & 54.47 & 56.96 & 59.10 \\
  &   & \checkmark & \checkmark & 60.36 & 61.13 & 55.92 & 56.21 & 57.76 & 60.85 \\
  & \checkmark & \checkmark & \checkmark & \underline{60.56
  }& \underline{61.62} & \underline{57.09} & \underline{56.25} & \underline{59.57} & \underline{60.94} \\
\checkmark & \checkmark & \checkmark & \checkmark & \textbf{62.80} & \textbf{62.84} & \textbf{61.75} & \textbf{64.03} & \textbf{62.51} & \textbf{62.80} \\
\hline
\end{tabular}}
\label{tab:enocder_ablation}
\end{table*}

\begin{table*}[t]
\centering
\caption{Ablation study of the Hierarchical Reasoning Decoder on the DRSeg benchmark. Each configuration progressively increases the decoder depth from two to three layers. Metrics (\%) are reported as gIoU~$\uparrow$ and cIoU~$\uparrow$.}
\setlength{\tabcolsep}{6pt}
\renewcommand{\arraystretch}{1.15}
\resizebox{0.85\textwidth}{!}{
\begin{tabular}{l|cc|cc|cc|cc}
\hline
\multirow{2}{*}{\textbf{Decoder Layer}} &
\multicolumn{2}{c|}{\textbf{Attribute Reasoning}} &
\multicolumn{2}{c|}{\textbf{Scene Reasoning}} &
\multicolumn{2}{c|}{\textbf{Spatial Reasoning}} &
\multicolumn{2}{c}{\textbf{ALL}} \\ 
\cline{2-9}
 & $gIoU\uparrow$ & $cIoU\uparrow$ & $gIoU\uparrow$ & $cIoU\uparrow$ & $gIoU\uparrow$ & $cIoU\uparrow$ & $gIoU\uparrow$ & $cIoU\uparrow$ \\ 
\hline
2-layers & 56.27 & 57.23 & 55.13 & 56.28 & 56.67 & 61.42 & 56.43 & 58.01 \\
\textbf{3-layers (Ours)} & \textbf{62.80} & \textbf{62.84} & \textbf{61.75} & \textbf{64.03} & \textbf{62.51} & \textbf{62.80} & \textbf{62.35} & \textbf{63.23} \\
\hline
\end{tabular}}
\label{tab:decoder_ablation}
\end{table*}

Following LLaVA-HR~\cite{luo2024feast, liu2024improved}, PixDLM is built on a simple and effective idea:  
\emph{inject high-resolution structure into low-resolution semantic reasoning through MultiPath  Alignment, allowing an MLLM to perform pixel-level segmentation without increasing visual token cost.} The architecture consists of three tightly coupled components:

(1) A \textbf{Dual-Path Vision Encoder} that separately models global semantics and fine structural details, then aligns and fuses them through a hierarchical cross-path mechanism.

(2) An \textbf{LLM backbone} that receives text tokens, image tokens, and a learnable \textit{Mask Token}~\cite{cheng2022masked}, whose hidden states naturally encode multimodal reasoning~\cite{lin2024video}.

(3) A \textbf{Hierarchical Reasoning Decoder} that progressively reconstructs the segmentation mask through cross-scale refinement using both high-resolution visual features and LLM-driven mask embeddings.

During inference, the image and instruction are processed by the LLM, which outputs the hidden state of the Mask Token containing integrated semantic+visual reasoning. To preserve spatial fidelity, PixDLM augments this reasoning signal with the final three high-resolution layers from the SAM encoder~\cite{KirillovMRMRGXW23}. These multi-scale features and mask embeddings are then decoded into the final mask~\cite{ronneberger2015u}. The overall pipeline is shown in Fig.~\ref{fig:method}.

Below, we detail each module and naturally introduce the mathematical formulation supporting these components.

\subsection{Dual-Path Vision Encoder}

UAV imagery requires jointly modeling (1) long-range context for reasoning, and (2) high-frequency boundaries for small-object segmentation. To satisfy these competing needs, PixDLM adopts a \textbf{Dual-Path Vision Encoder} consisting of:
\begin{itemize}
    \item \textbf{Global-Scope Path}: Captures coarse semantics for long-range reasoning.
    \item \textbf{Fine-Scope Path}: Preserves dense structural details for small-object boundaries.
\end{itemize}
Overall, the proposed Dual-Path Vision Encoder follows a simple three-step pipeline: \emph{(1) dual-path feature extraction, (2) three-stage latent fusion, and (3) final fusion.}

To jointly preserve global semantics and fine-grained structures, we employ a \textbf{dual-path feature extraction} strategy, where a dual-path visual encoder consisting of a Global-Scope Path and a Fine-Scope Path. Given an input image $x$, the two branches operate on different resolutions:
\begin{equation}
F_f^{(0)} = E_f(R_f(x)), \qquad F_s = E_s(R_s(x)),
\end{equation}
where $R_f(\cdot)$ and $R_s(\cdot)$ denote branch-specific resizing operations, and $E_f$ and $E_s$ denote the CLIP~\cite{radford2021learning} and SAM encoders~\cite{KirillovMRMRGXW23}, respectively. Here, $F_f^{(0)}$ represents the initial token features from the Global-Scope Path, while $F_s$ denotes the high-resolution structural feature extracted by the Fine-Scope Path.



To bridge the resolution gap, we introduce a lightweight \textbf{MultiPath Alignment}. Rather than fusing the two branches only at the output level, we inject the feature from the Fine-Scope Path into the intermediate latent spaces of the Global-Scope Path in \textbf{three-stage latent fusion}. Specifically, at each stage $k \in \{1,2,3\}$, we first align the  feature from Fine-Scope Path to the token dimension and spatial resolution of the current Global-Scope Path latent:
\begin{equation}
\tilde{F}_s^{(k)} = A_k(F_s), \qquad k \in \{1,2,3\},
\end{equation}
where $A_k(\cdot)$ denotes the alignment operator at stage $k$, including channel projection and spatial resampling. In this way, $\tilde{F}_s^{(k)}$ is compatible with the latent feature of the corresponding CLIP stage.

After alignment, we perform gated residual fusion:
\begin{equation}
F_f^{(k)} = B_k(F_f^{(k-1)}) + G_k\!\left(B_k(F_f^{(k-1)}), \tilde{F}_s^{(k)}\right)\odot \tilde{F}_s^{(k)},
\end{equation}
where $B_k(\cdot)$ denotes the transformation of the $k$-th Global-Scope Path stage, $G_k(\cdot)$ is a learnable gating function, and $\odot$ denotes element-wise scaling. This design allows the model to adaptively inject high-resolution structural cues into the semantic encoding process, so that fine details can influence representation learning throughout the encoder rather than only at the end.
\textbf{Final Fusion:}
After the three-stage latent fusion, we further perform an explicit fusion at the output level. The final feature from Global-Scope Path and the aligned  feature from Fine-Scope Path are projected into the same feature space and summed to form the final visual representation:
\begin{equation}
F_{\mathrm{out}} = W_f F_f^{(3)} + W_s A_o(F_s),
\end{equation}
where $A_o(\cdot)$ denotes the output-level alignment operator, and $W_f$ and $W_s$ are learnable linear projections for the global and fine scope path, respectively. The resulting feature $F_{\mathrm{out}}$ is then used as the final visual representation for downstream language reasoning and mask prediction.

\subsection{Hierarchical Reasoning Decoder}

The Mask Token produced by the LLM contains cross-modal reasoning signals
but lacks fine spatial granularity.
To recover pixel-level masks, we introduce a lightweight 
\textbf{Hierarchical Reasoning Decoder}
that progressively injects multi-scale visual features into the reasoning flow.

At each layer $l$, the decoder receives:
(1) visual feature map $\mathbf{F}^l$, and  
(2) a higher-level predicted mask $M^{l+1}$.
The mask modulates feature importance through:
\begin{equation}
\mathbf{F}'^l = \mathbf{F}^l \odot (\sigma(M^{l+1}) + 1),
\end{equation}
directing attention to high-confidence regions while suppressing noise.

After obtaining all $n$ intermediate masks $\{M^1,\dots,M^n\}$,
we fuse them using learnable layer-wise coefficients:
\begin{equation}
\hat{M} = \sum_{l=1}^{n} \gamma^l M^l.
\end{equation}
This hierarchical refinement improves spatial precision and enforces semantic consistency across scales, which is critical for UAV reasoning segmentation where objects vary drastically in size and appearance.

%% file: sec/5_experiment.tex
\section{Experiment}
\subsection{Setting}
We initialize our \textbf{VLLM} with the pre-trained weights of \textbf{LLaMA-2-13B}~\cite{abs-2307-09288} as the language backbone and adopt a \textbf{Dual-Path vision encoder} for multi-scale feature extraction. 
Training is performed with \textbf{DeepSpeed ZeRO-2}~\cite{RajbhandariRRH20} using a batch size of 2 and 10 gradient-accumulation steps on eight NVIDIA GeForce RTX~3090 GPUs. 
We employ the \textbf{AdamW} optimizer with a learning rate of $3\times10^{-4}$, a linear warm-up of 100 steps, and train for 10 epochs (200 steps per epoch). 
The total loss is a weighted combination of cross-entropy~(1.0), binary cross-entropy~(2.0), and Dice loss~(0.5).

\subsection{Comparisons with State-of-the-Arts}
Table~\ref{tab:drseg_results} compares PixDLM with state-of-the-art approaches.Fig.~\ref{fig:comparison} further provides visual comparisons of segmentation results produced by LISA~\cite{LaiTCLY0J24}, our fine-tuned LISA variant, GeoPix~\cite{abs-2501-06828}, and our PixDLM model.
In the zero-shot setting, existing models show limited performance on DRSeg, with low gIoU and cIoU across all three reasoning dimensions. This suggests that current multimodal models do not transfer well to UAV reasoning segmentation, underscoring the need for a dataset specifically designed for this setting.

With supervised fine-tuning (SFT), our method achieves higher scores than baselines such as PixelLM across all reasoning dimensions, demonstrating a clear performance advantage. These results indicate that PixDLM aligns more effectively with the demands of DRSeg and provides stronger reasoning and segmentation capabilities, validating the effectiveness of our design.Fig.~\ref{fig:segmentation results} further visualizes PixDLM’s segmentation results across the three reasoning subsets of DRSeg.

To assess its cross-domain generalization, we further evaluate PixDLM on the RefCOCOs~\cite{KazemzadehOMB14}, RefCOCO+~\cite{yu2016modeling}, and RefCOCOg~\cite{mao2016generation} dataset (See in the Supplementary Material).
PixDLM delivers strong performance across all three benchmarks and demonstrates clear advantages over multimodal baselines such as LISA-7B and PixelLM on most evaluation splits.
These findings suggest that the proposed Dual-Path architecture and reasoning-aware alignment not only enhance UAV segmentation, but also transfer effectively to standard referring segmentation tasks, underscoring the model’s robust generalization capability.

\subsection{Ablations}
\textbf{Training Strategy} (with/without COT; mixed vs single-task)
As shown in Tab~\ref{tab:cot_ablation}, training purely on DRSeg provides a strong baseline (Attribute: $62.51\%$~gIoU / $61.62\%$~cIoU; Scene: $55.92\%$ / $54.29\%$; Spatial: $59.57\%$ / $60.94\%$). 
Adding COT supervision yields the best overall configuration, improving to $62.80\%$ / $62.84\%$ on Attribute and $61.75\%$ / $64.03\%$ on Scene, while maintaining solid Spatial performance ($62.51\%$ / $62.80\%$). 
Using RRSIS-D alone with DRSeg brings limited gains, likely due to weaker reasoning annotations. 
Overall, these results show that (i) COT provides the largest overall benefit by injecting step-wise linguistic reasoning signals, (ii) single-task training underperforms mixed-task learning across dimensions.

\textbf{MultiPath Alignment Layers}
Tab~\ref{tab:enocder_ablation} examines the four fusion layers in our Dual-Path Vision Encoder. 
Starting with no fusion (all columns empty), the model attains $58.39\%$/$57.35\%$ (Attribute), $55.19\%$/$54.29\%$ (Scene), and $55.38\%$/$54.47\%$ (Spatial). 
Enabling fusion progressively from deep to shallow yields monotonic gains across all reasoning types. 
Activating the full set of 1st--4th layers (i.e., all four fusion stages) delivers the best results: $62.80\%$/$62.84\%$ (Attribute), $61.75\%$/$64.03\%$ (Scene), and $62.51\%$/$62.80\%$ (Spatial). 
Relative to no fusion, this corresponds to $+4.41\%$/$+5.49\%$ on Attribute, $+6.56\%$/$+9.74\%$ on Scene, and $+7.13\%$/$+8.33\%$ on Spatial. 
These results confirm the importance of MultiPath Alignment across scales: early fusion introduces structural details, deeper fusion stabilizes semantics, and combining all four stages achieves the most consistent gains.

\textbf{Hierarchical Reasoning Decoder}
As summarized in Table~\ref{tab:decoder_ablation}, increasing the decoder depth from two layers (two-scale fusion) to three layers (three-scale fusion) leads to clear and consistent performance gains. The overall average increases from $56.43\%$/$58.01\%$ to $62.35\%$/$63.23\%$ in gIoU/cIoU, corresponding to improvements of $+5.92\%$ and $+5.22\%$, respectively. 
Performance also improves across all reasoning dimensions. For Attribute Reasoning, the score increases from $56.27\%$/$57.23\%$ to $62.80\%$/$62.84\%$; for Scene Reasoning, from $55.13\%$/$56.28\%$ to $61.75\%$/$64.04\%$; and for Spatial Reasoning, from $56.67\%$/$61.42\%$ to $62.51\%$/$62.80\%$. 
These results show that the three-scale decoder improves coarse-to-fine refinement and multimodal integration during mask prediction, achieving a good balance between accuracy and efficiency. Therefore, we use the three-layer decoder as the default configuration.

%% file: sec/6_conclusion.tex
\section{Conclusion}
We introduce UAV Reasoning Segmentation and categorize reasoning into three types: attribute, scene, and spatial reasoning, which collectively characterize the requirements of UAV vision scenarios.
To support this task, we construct DRSeg, a high-resolution, multi-view UAV dataset annotated with detailed chain-of-thought supervision.
We further propose PixDLM, a multimodal model that integrates a Dual-Path encoder with a hierarchical decoder.
Experimental results demonstrate that PixDLM achieves substantial improvements in both semantic consistency and spatial localization.

%% file: sec/7_acknowledgments.tex
\section{Acknowledgments}
This work is supported by the National Key Research and Development Program of China (No. 2025YFE0113500), National Science Fund for Distinguished Young Scholars (No.62525605) and the National Natural Science Foundation of China (No. U25B2066, No. U22B2051, No. 62302411).

%% file: sec/9_supple.tex
\clearpage
\setcounter{page}{1}
\maketitlesupplementary

In this supplementary material, Sec.~\ref{sec:count} provides additional visualizations of the DRSeg dataset. Sec.~\ref{sec:B} discusses the design of the loss functions. Sec.~\ref{sec:refcoco} presents extended experimental results, covering the baselines of GeoPix and Seg-Zero across the three reasoning dimensions, along with the performance of PixDLM on the ReferSeg benchmark. Sec.~\ref{sec:D} presents ablation studies on alternative Multi-Path Encoder fusion strategies and a one-layer decoder variant.

\renewcommand{\thesection}{\Alph{section}}
\renewcommand{\thetable}{\thesection.\arabic{table}}
\renewcommand{\thefigure}{\thesection.\arabic{figure}}
\setcounter{section}{0}
\setcounter{table}{0}
\setcounter{figure}{0}

\section{DRSeg}
\label{sec:count}
\subsection{Prompt Design}
\begin{figure*}[t]
    \centering
    \includegraphics[width=\linewidth]{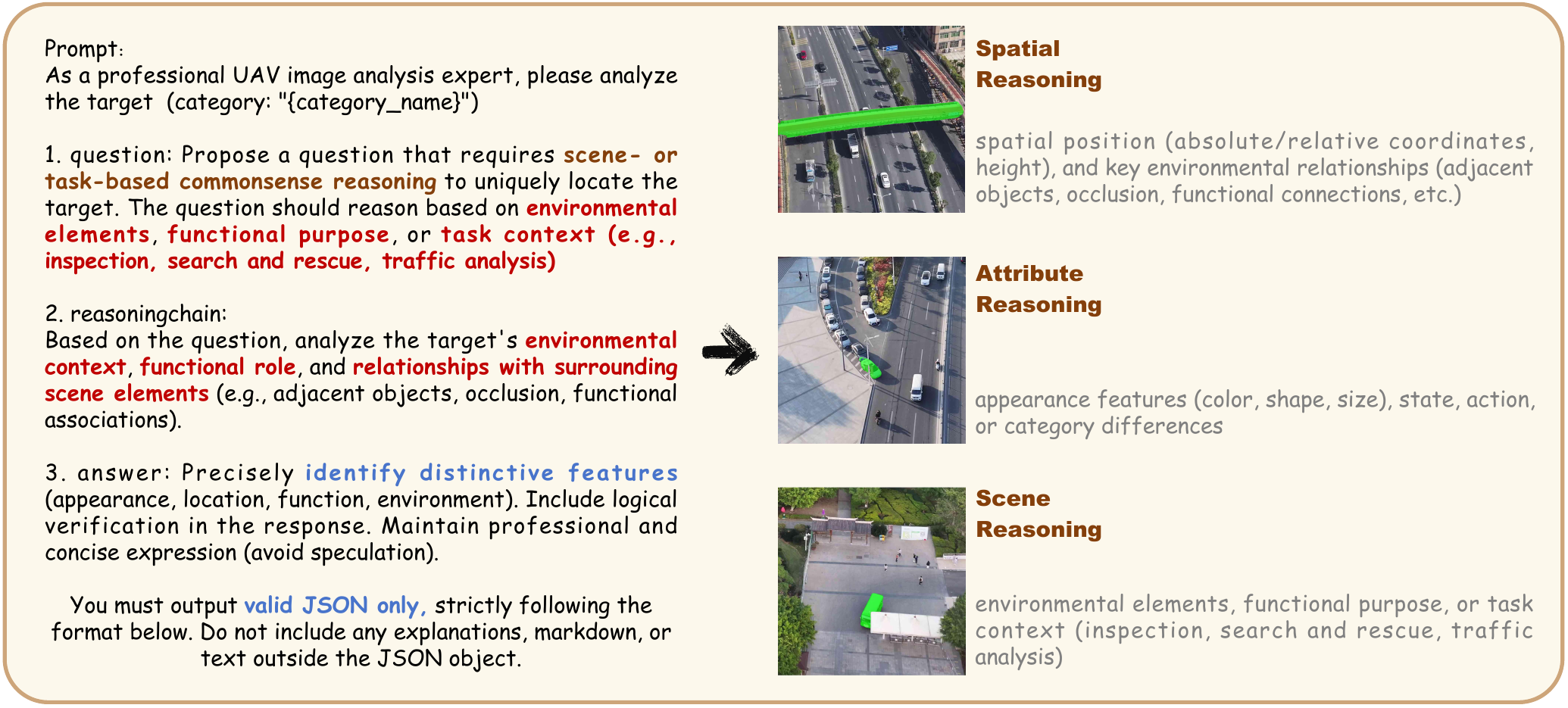}
    \caption{Instruction template used for constructing reasoning annotations in DRSeg. 
    It includes structured prompts for generating Spatial, Attribute, and Scene-level reasoning, each guiding the model to produce questions, reasoning chains, and answers in a unified JSON format.}
    \label{fig:prompt}
\end{figure*}

Figure~\ref{fig:prompt} illustrates the instruction template used to construct the reasoning annotations in DRSeg. To ensure consistent and high-quality supervision,we design a unified prompt format that guides the model to generate three types of reasoning signals: Spatial, Attribute, and Scene-level reasoning. 
For each annotated instance, the prompt requires: 
(1) a reasoning-oriented question that uniquely locates the target; 
(2) a structured reasoning chain that analyzes contextual cues such as environmental elements, functional roles, spatial relations, and appearance features; and 
(3) a concise, verifiable answer. 
The output is constrained to a strict JSON format, enabling robust automatic parsing and ensuring annotation consistency across all 10,000 samples.
\subsection{Dataset Statistics}
\begin{figure*}[t]
    \centering
    \includegraphics[width=\linewidth]{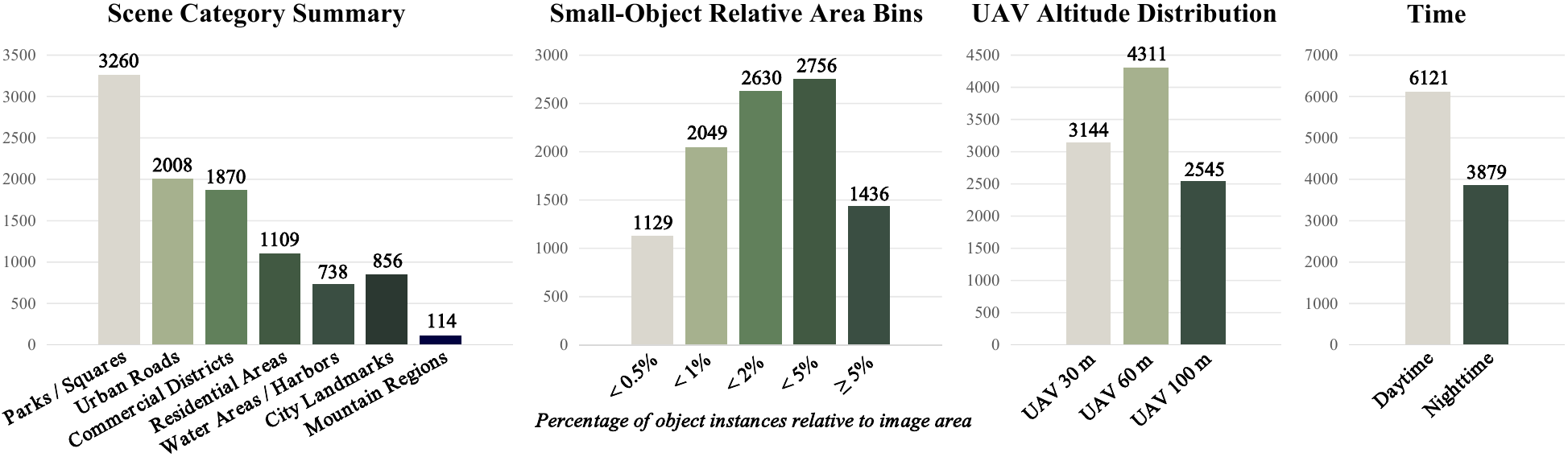}
    \caption{Dataset statistics of DRSeg, including small-object relative area distribution, 
    scene category summary, UAV altitude distribution, and day/night proportions.}
    \label{fig:dataset_stats}
\end{figure*}

We provide extended visualizations of DRSeg's key statistics in Table~\ref{fig:dataset_stats}, including scene category distribution, altitude levels, day/night proportions, and object scale diversity. These plots offer a more comprehensive view of the dataset's geometric, semantic, and environmental variability.

\subsection{Human Verification of DRSeg Annotations}
To systematically assess the annotation quality of \textbf{DRSeg}, we invited remote sensing researchers to conduct a human verification study on a stratified random sample of 2,000 images. 
The evaluation covers three dimensions: reasoning question uniqueness, CoT reasoning consistency, and Mask--text alignment. Each sample was independently reviewed and quantified based on the following criteria.

\textbf{Uniquely Identifiable Rate}: This metric measures whether a reasoning question can uniquely refer to a single target entity in the image. We compute the proportion of samples for which both annotators agree that the question is uniquely identifiable.

\textbf{CoT Reasoning Consistency}: This metric evaluates whether the chain-of-thought reasoning is consistent with the visual evidence and the final target. Each CoT explanation is scored according to the following three criteria, with $1$ point awarded for each satisfied condition (score range: $0$--$3$):
    \begin{itemize}
        \item The entities mentioned in the CoT exist in the image;
        \item The reasoning steps clearly describe the localization process;
        \item The final predicted target matches the image semantics.
    \end{itemize}

    The average CoT score is defined as:
    \[
    \text{CoT-Score} = \frac{1}{N} \sum_{i=1}^{N} s_i,
    \]
    where $s_i$ denotes the CoT score of the $i$-th sample.

\textbf{MaskAlign-Rate}: This metric measures whether the segmentation mask aligns with the textual description. A sample is considered aligned if (1) the target region coverage is $\geq 95\%$, and (2) the non-target miscoverage is $\leq 5\%$. We report the proportion of samples for which both annotators judged the mask to be aligned.

The results show that the Uniquely Identifiable Rate reaches \textbf{95\%}, the CoT-Score achieves \textbf{2.83} out of $3$, and the MaskAlign-Rate is \textbf{97\%}. These findings demonstrate that DRSeg exhibits highly reliable annotation quality in terms of reasoning target identification, reasoning-chain soundness, and visual-text alignment.

\begin{table*}[t]
\centering
\caption{Zero-shot comparison on the \textbf{DRSeg} benchmark across three reasoning dimensions.}
\large  
\setlength{\tabcolsep}{6pt}
\renewcommand{\arraystretch}{1.15}
\resizebox{0.9\textwidth}{!}{
\begin{tabular}{l|l|cc|cc|cc}
\hline
\multirow{2}{*}{\textbf{Setting}} &
\multirow{2}{*}{\textbf{Model Name}} &
\multicolumn{2}{c|}{\textbf{Attribute Reasoning}} &
\multicolumn{2}{c|}{\textbf{Scene Reasoning}} &
\multicolumn{2}{c}{\textbf{Spatial Reasoning}} \\ 
\cline{3-8}
 & & $gIoU\uparrow$ & $cIoU\uparrow$ & $gIoU\uparrow$ & $cIoU\uparrow$ & $gIoU\uparrow$ & $cIoU\uparrow$ \\ 
\hline
\multirow{2}{*}{Zero-shot}
& Seg-Zero-7B~\cite{abs-2503-06520}     & 49.32 & 50.58 & 25.99 & 29.56 & 31.12 & 36.25 \\
& GeoPix-7B~\cite{abs-2501-06828}         & 42.96 & 46.51 & 36.84 & 41.67 & 36.79 & 39.73 \\
\hline
\end{tabular}}
\label{tab:drseg_zeroshot}
\end{table*}
\begin{table*}[t]
\centering
\caption{Results on the referring segmentation benchmark.} 
\setlength{\tabcolsep}{6pt}
\renewcommand{\arraystretch}{1.15}
\resizebox{0.7\textwidth}{!}{
\begin{tabular}{l|ccc|ccc|cc}
\hline
\multirow{2}{*}{\textbf{Model Name}} &
\multicolumn{3}{c|}{\textbf{refCOCO}} &
\multicolumn{3}{c|}{\textbf{refCOCO+}} &
\multicolumn{2}{c}{\textbf{refCOCOg}} \\
\cline{2-9}
 & val & testA & testB & val & testA & testB & val & test \\
\hline
GSVA-7B~\cite{abs-2312-10103} & \underline{76.4} & 77.4 & \underline{72.8} & 64.5 & 67.7 & 58.6 & \underline{71.1} & \underline{72.0} \\
LaSagnA-7B~\cite{abs-2404-08506} & \textbf{76.8} & \underline{78.7} & \textbf{73.8} & \underline{66.4} & 70.6 & \underline{60.1} & 70.6 & 71.9 \\
LISA-7B~\cite{LaiTCLY0J24} & 74.1 & 76.5 & 71.1 & 62.4 & 67.4 & 56.5 & 64.5 & 66.7 \\
PixelLM~\cite{RenHW0FFJ24} & 73.0 & 76.5 & 68.2 & 66.3 & \underline{71.7} & 58.3 & 69.3 & 70.5 \\
\textbf{Ours} & 75.2 & \textbf{80.2} & 70.5 & \textbf{68.5} & \textbf{73.3} & \textbf{60.6} & \textbf{73.3} & \textbf{72.8} \\
\hline
\end{tabular}}
\label{tab:refcoco_results}
\end{table*}

\section{Training Configuration }
\label{sec:B}
\begin{table*}[t]
\centering
\caption{
Ablation study of the \textbf{Multi-Path Alignment Fusion Strategy} on the \textbf{DRSeg} benchmark.
We compare three fusion directions: \textbf{SAM$\rightarrow$ CLIP}, \textbf{SAM+CLIP}, and \textbf{CLIP$\rightarrow$ SAM}.
Metrics (\%) are reported as $gIoU\uparrow$ and $cIoU\uparrow$.
}
\setlength{\tabcolsep}{6pt}
\renewcommand{\arraystretch}{1.15}
\resizebox{0.8\textwidth}{!}{
\begin{tabular}{c|cc|cc|cc}
\hline
\multirow{2}{*}{\textbf{Fusion Strategy}} &
\multicolumn{2}{c|}{\textbf{Attribute Reasoning}} &
\multicolumn{2}{c|}{\textbf{Scene Reasoning}} &
\multicolumn{2}{c}{\textbf{Spatial Reasoning}} \\
\cline{2-7}
 & $gIoU\uparrow$ & $cIoU\uparrow$ 
 & $gIoU\uparrow$ & $cIoU\uparrow$
 & $gIoU\uparrow$ & $cIoU\uparrow$ \\
\hline

SAM + CLIP 
& 60.11 & 58.73
& 56.80 & 56.23
& 59.34 & 60.32 \\

CLIP $\rightarrow$ SAM
& \underline{60.24} & \underline{60.28}
& \underline{57.12} & \underline{57.84}
& \underline{59.36} & \underline{60.97} \\

SAM $\rightarrow$ CLIP 
& \textbf{62.80} & \textbf{62.84}
& \textbf{61.75} & \textbf{64.03}
& \textbf{62.51} & \textbf{62.80} \\

\hline
\end{tabular}}
\label{tab:multipath_fusion}
\end{table*}

\begin{table*}[t]
\centering
\caption{Ablation study of single-layer decoders on the DRSeg benchmark.
We compare our Single-Layer Decoder (no fusion) with the SAM Decoder baseline.Metrics (\%) are reported as gIoU~$\uparrow$ and cIoU~$\uparrow$.}
\setlength{\tabcolsep}{6pt}
\renewcommand{\arraystretch}{1.15}
\resizebox{0.8\textwidth}{!}{
\begin{tabular}{l|cc|cc|cc}
\hline
\multirow{2}{*}{\textbf{Decoder Layer}} &
\multicolumn{2}{c|}{\textbf{Attribute Reasoning}} &
\multicolumn{2}{c|}{\textbf{Scene Reasoning}} &
\multicolumn{2}{c}{\textbf{Spatial Reasoning}} \\ 
\cline{2-7}
 & $gIoU\uparrow$ & $cIoU\uparrow$ & $gIoU\uparrow$ & $cIoU\uparrow$ & $gIoU\uparrow$ & $cIoU\uparrow$ \\ 
\hline
1-layer & 47.06 & 50.07 & 45.16 & 51.35 & 45.81 & 53.27 \\
sam2.1-l& 46.43 & 48.80 & 44.78 & 49.90 & 46.16 & 51.95 \\
\hline
\end{tabular}}
\label{tab:decoder_single_layer}
\end{table*}

\subsection{Loss Design}
The overall objective $\mathcal{L}$ constitutes the weighted sum of these losses, calibrated by $\lambda_{\text{ref}}$ and $\lambda_{\text{dice}}$:
\begin{equation}
\mathcal{L}
=
\mathcal{L}_{\text{txt}}
+
\lambda_{\text{ref}}\, \mathcal{L}_{\text{ref}}
+
\lambda_{\text{dice}}\, \mathcal{L}_{\text{dice}} .
\end{equation}
$L$ denotes the overall training objective, $L_{\text{txt}}$ is the text cross-entropy loss, 
$L_{\text{ref}}$ is the binary cross-entropy loss supervising the referred target mask, 
and $L_{\text{dice}}$ is the Dice loss measuring region-level overlap. 
Following LISA~\cite{LaiTCLY0J24}, PixDLM keeps the same CE and Dice loss weights 
($\lambda_{\text{ref}} = 2.0$ and $\lambda_{\text{dice}} = 0.5$) 
\subsection{Trainable Parameters}
To preserve the knowledge of the pre-trained multimodal LLM, PixDLM adopts LoRA for efficient fine-tuning while completely freezing the CLIP vision encoder, the SAM vision encoder, and the base LLaMA/LLaVA language backbone (only the inserted LoRA adapters carry gradients). The trainable components include the MultiPath Alignment modules, the hierarchical reasoning decoder, the text-hidden projection MLP, the SAM-to-LLM bridging convolution , the prompt encoder, and the image-feature neck adapter.
\subsection{Inference Efficiency and Trainable Parameter Budget}
Baseline evaluations on an RTX 3090 GPU show that PixDLM requires an average of 1.12 seconds ($\approx$0.89 FPS) to process a single $1024 \times 1024$ UAV image. This measurement is obtained under the full model configuration, including the multi-level segmentation decoder and the dual-path semantic–structural alignment mechanism. The results indicate that PixDLM maintains stable throughput even with fine-grained structural modeling and cross-path interaction enabled, making it suitable for offline analysis and large-scale batch inference. In addition, the model contains 4,194,304 trainable parameters ($\approx$4.19M) out of a total of 7,303,624,675 parameters ($\approx$7.3B), corresponding to a trainable ratio of 0.0574\%.

\setcounter{figure}{0}
\setcounter{table}{0}

\renewcommand{\thefigure}{C.\arabic{figure}}
\renewcommand{\thetable}{C.\arabic{table}}

\section{Extended Experimental Results}
\label{sec:refcoco}
\subsection{Additional SOTA Comparisons}
To further strengthen the comprehensiveness of our evaluation, we additionally benchmark two recently proposed multimodal segmentation models \textbf{Seg-Zero-7B}~\cite{abs-2503-06520} and \textbf{GeoPix-7B}~\cite{abs-2501-06828} under the zero-shot setting in Table~\ref{tab:drseg_zeroshot}
These models have demonstrated strong generalization capabilities in open-world or geospatial reasoning tasks, making them relevant candidates for comparison on the DRSeg benchmark.

\subsection{ReferSeg Results}
Table~\ref{tab:refcoco_results} shows that PixDLM, despite being designed 
primarily for UAV reasoning segmentation, transfers well to standard referring 
segmentation tasks. The model achieves leading accuracy on multiple splits of 
refCOCO and refCOCO+, and remains highly competitive on refCOCOg. 
These results highlight the versatility and robustness of our dual-path design and Multi-Path Alignment module.

\section{Ablation Studies}
\label{sec:D}
\subsection{Ablations on Multi-Path Fusion Schemes}
Table~\ref{tab:multipath_fusion} reports an ablation study on the fusion
directions of the Multi-Path Alignment module. We compare three alternatives:
(\emph{i}) directly summing features from both paths (SAM+CLIP),
(\emph{ii}) injecting SAM features into the CLIP branch (CLIP$\rightarrow$SAM),
and (\emph{iii}) injecting CLIP features into the SAM branch (SAM$\rightarrow$CLIP).
The results show that the SAM$\rightarrow$CLIP direction consistently yields the
best performance across all three reasoning dimensions. This confirms that
guiding the low-resolution semantic path with high-resolution structural cues is
more effective than the reverse integration strategy.

\subsection{Ablations on Hierarchical Reasoning Decoder}
To better understand the contribution of hierarchical multi-level reasoning, we conduct ablations on non-hierarchical decoder variants, as shown in Table~\ref{tab:decoder_single_layer}. Specifically, we compare two single-layer decoders: (1) a \textbf{Single-Layer Decoder} that performs one-step reasoning without any cross-level fusion, and (2) a \textbf{SAM2.1-L Decoder} baseline that replaces our decoder with the mask prediction head from SAM2.1-Hiera-L.

Both variants operate without hierarchical fusion, thus isolating the effect of multi-stage reasoning depth. The results demonstrate that all single-layer designs exhibit a clear performance drop across attribute, scene, and spatial reasoning dimensions. In particular, our Single-Layer Decoder achieves higher scores than the SAM decoder baseline, indicating that even without hierarchical fusion, a reasoning-oriented decoder provides better cross-modality alignment than a purely segmentation-oriented head.

However, both single-layer variants lag significantly behind our full hierarchical decoder (Table 4 in the main paper), highlighting the importance of progressive multi-level reasoning for capturing complex UAV semantics, long-range relations, and fine-grained spatial cues. These results verify that hierarchical fusion is essential for robust reasoning segmentation under UAV-specific challenges.